# Physics-guided machine learning predicts the planet-scale performance of solar farms with sparse, heterogeneous, public data

Jabir Bin Jahangir[a] and Muhammad Ashraful Alam[a]

[a]Department of Electrical and Computer Engineering, Purdue University, West Lafayette, IN 47906

The photovoltaics (PV) technology landscape is evolving rapidly. To predict the potential and scalability of emerging PV technologies, a global understanding of these systems' performance is essential. Traditionally, experimental and computational studies at large national research facilities have focused on PV performance in specific regional climates. However, synthesizing these regional studies to understand the worldwide performance potential has proven difficult. Given the expense of obtaining experimental data, the challenge of coordinating experiments at national labs across a politically-divided world, and the data-privacy concerns of large commercial operators, however, a fundamentally different, data-efficient approach is desired. Here, we present a physics-guided machine learning (PGML) scheme to demonstrate that: (a) The world can be divided into a few PV-specific climate zones, called PVZones, illustrating that the relevant meteorological conditions are shared across continents; (b) by exploiting the climatic similarities, high-quality monthly energy yield data from as few as five locations can accurately predict yearly energy yield potential with high spatial resolution and a root mean square error of less than 8 kW h m$^{-2}$, and (c) even with noisy, heterogeneous public PV performance data, the global energy yield can be predicted with less than 6% relative error compared to physics-based simulations provided that the dataset is representative. This PGML scheme is agnostic to PV technology and farm topology, making it adaptable to new PV technologies or farm configurations. The results encourage physics-guided, data-driven collaboration among national policymakers and research organizations to build efficient decision support systems for accelerated PV qualification and deployment across the world.

photovoltaics| solar farm | machine learning | physics-guided | energy yield

The fast-paced advancement of material platforms such as silicon, perovskite, and CdTe, along with developments in cell and module technologies, such as PERC, PERT, bifacial, tandem, etc., alongside various farm layouts such as tilted, tracking, vertical, floating, and agro-PV, has generated a pressing demand to assess the energy yield (EY) and economic feasibility of different technological choices and farm arrangements across diverse global regions. After all, modern solar farms require significant investment, and incorrect decisions have long-term financial, political, and public perception implications. A number of free and commercial and open-source physics-based model exist for solar farm performance prediction, e.g., PVSyst, Bifacial Radiance, PlantPredict, PVMAPS. These models are widely used and have been critically important for planning and decision making. Currently, predicting EY of solar farms is supported either by such time-intensive computational modeling involving proprietary software operated by modeling experts from various consulting farms and/or by integrating the scattered field reports from highly-instrumented test centers located at a few technologically-advanced countries in the world. An alternative framework that allows fast model prediction and rapid experimental validation leveraging machine-learning (ML) techniques would transform the development and deployment of solar renewables by allowing independent projections of PV performance and their validation across the world.

Unfortunately, a purely data-driven ML approach requires voluminous field data, yet they may produce results that are inconsistent with physical laws (1, 2). Indeed, a significant amount of PV performance data has been published in the literature and are available from public data dashboards from PV systems deployed at airports, universities, and weather-stations. However, since the combined data reflects a melange of materials, module types, tilt-angles, farm configurations, climate conditions, and time granularity, it is unclear how to address the issues of heterogeneity in such datasets to train a reliable ML model. By focusing on the physical understanding of the data and process being modeled, a physics-guided machine-learning (PGML) approach reduces data dependency and improves prediction accuracy. As such, PGML could transform the decision-making process by significantly accelerating the analysis and design time by orders of magnitude. Recognizing this potential, PGML approaches are being explored at all system scales ranging from cell to farm, including material discovery(3), solar cell optimization (4, 5), solar resource prediction (6), module degradation mode identification(7). and solar farm performance prediction (4, 8).

Previously, Patel et al. (8) have demonstrated that an ML-based functional approximation guided by the physics of solar farms can reduce the prediction time by orders of magnitude (i.e., from hours to seconds). Remarkably, the worldwide energy yield potential can can be predicted with less than 10% RMSE with a relatively small dataset ($N = 100$) of simulated yearly yield observations from randomly chosen locations across the world. This suggested the possibility of a data-driven approach to PV yield prediction based on high-quality experimental data. Despite the impressive results, however, a manufacturer may find it impractical or expensive to install newly developed modules to collect year-long yield data from hundreds of locations across the globe. Therefore, while the scheme proposed in (8) addressed the challenge of rapid prediction with a small dataset, the *cost-effective, field-validation* of the predictions remains an open question. In the context of PV, data efficient learning is particularly desirable due to the limited availability of publicly accessible, high-quality, long-term, global-scale PV performance data covering the gamut of climatic conditions across the world.

In this work, we show how the physical understanding of PV systems can inform the design of data-efficient, generalizable ML models for PV performance prediction. Specifically, we show that the EY potential of solar farms across the world can be predicted with high accuracy ($\leq$ 5% error) at a high spatial resolution (0.5° × 0.5°) even when a model trained with geographi-cally sparse, heterogeneous, and public data sets. We systematically address the data quantity and quality pre-requisites to train global-scale model. Specifically, we will demonstrate that diverse sampling of physically relevant variables is the key to train such models. By dividing the world into PV-specific climate zones, we show how one can achieve diverse/importance sampling to train ML models with small, yet *representative* datasets. The power of our idea lies in the seeing the climatic similarity across the world: PV data collected from a site also represents conditions experienced elsewhere and thus predict the performance of many geographically distant locations. As an aside, the proposed methodology is problem-agnostic in the sense that any technology metric that depends on local climate (e.g. reliability of automotive electronics) can be analyzed by dividing the world in technology-aware climate zones and validated using publicly available heterogeneous field data.

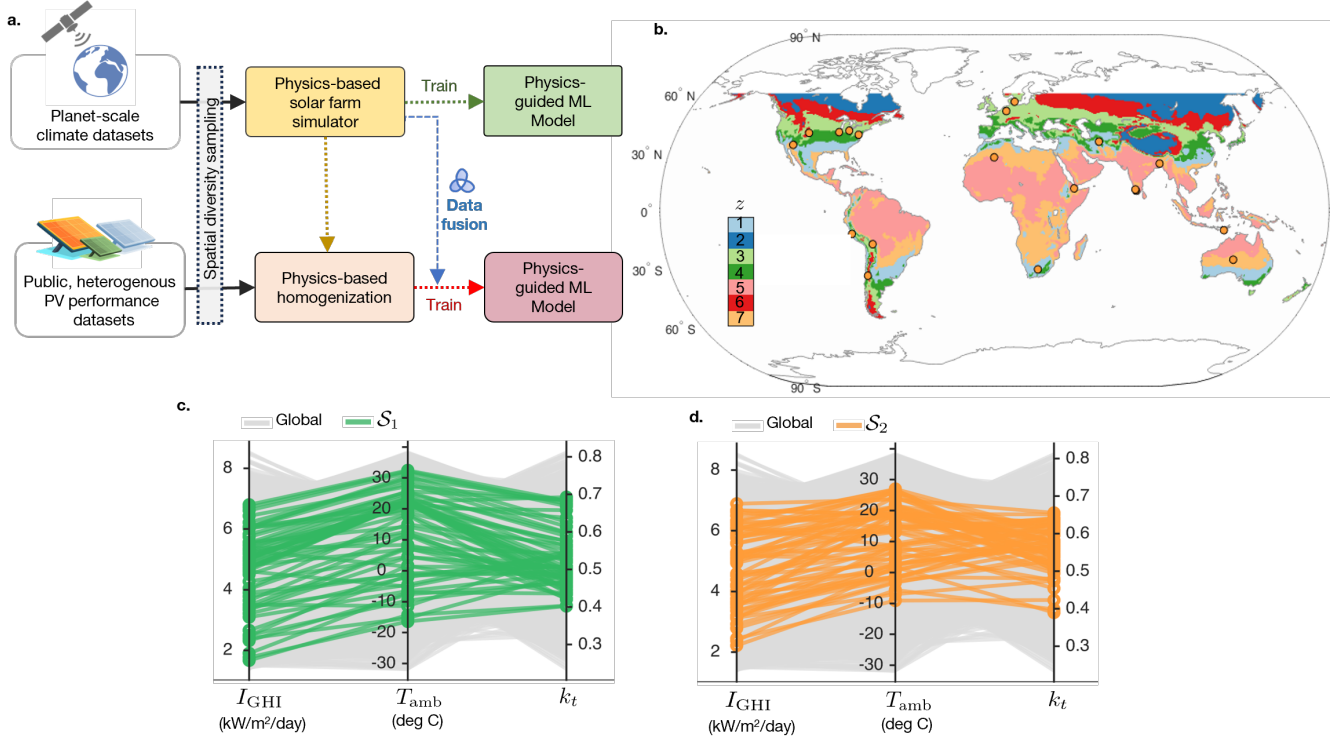

**Fig. 1.** (a) Overview of the PGML methodology for solar farms leveraging spatial diversity sampling. (b) PVZones (PVZ): geographical distribution of physically significant variables ($I_{\text{GHI}}$, $T_{\text{amb}}$, $k_t$) zoned with k-means clustering algorithm (k = 7). The map delineates the similarity between the features determining yield potential across continents that can be exploited to construct a training dataset with optimal set of test locations. (c-d) The representativeness of a dataset can be visualized by overlaying its observations (green and orange strings) on the global data (gray strings) in a parallel coordinates plot, which display the range of monthly average values for the variables observed worldwide.

## Using climate equivalence for data-efficient models

As mentioned, Patel et al.(8) demonstrated that one can reliably predict the annual energy yield of single-axis sun-tracking farms based on three variables, namely, location-specific average monthly global horizontal irradiance ($I_{\text{GHI}}$), clearness index ($k_t$), and ambient temperature ($T_{\text{amb}}$). In other words, the model implies that a specific combination of these input variables would produce identical energy yield.

This observation leads to four key ideas as shown in Fig. 1.

- *Spatial diversity sampling.* If the world can be segmented into regions defined by effectively similar input variables, then relatively few (functionally important) samples from each of these regions would train the a ML model far more efficiently than random sampling possibly can.

- *Temporal equivalence sampling.* The values of those variables seen at a location depend on the position of the sun, and thus, change monthly throughout the year. For example, West Lafayette feels like Arizona during the summer, and Alaska during the winter! Therefore, a ML model trained with monthly yield ($M$) data from a given location, could predict the yields at many other locations during different months.

- *Utilize heterogeneous public datasets.* The predictions of a PGML trained on large-scale synthetic data can be used to efficiently homogenize publicly available PV performance datasets. These datasets then can be used to train PGML model based on field data.

- *Physics-guided data fusion.* A large quantity of field data alone is insufficient to train a ML model that can generalize globally if there are gaps in representativeness. By combining field-collected data with synthetic that fill the gaps, one creates a more representative dataset suitable for training machine-learning models to predict global PV performance.

We will develop these key ideas in the next few sections.

## Physics-guided ML models for Solar Farms

Multiple approaches exist to develop a ML model guided by physics (9), with the general goal being to train a model that makes physically consistent predictions. One approach is to train the model with data obtained through numerical and/or physical experiments. This training data, consisting of physically-relevant inputs and corresponding outputs, embodies the physics of the data generating process. In this approach, an understanding of the physics underlying the modeled process is crucial as it directly guides the selection of variables. When the input domain of the problem is sufficiently represented in the training dataset, the model (e.g., Neural Network, Random Forest, etc.) is able to predict solutions for unknown inputs. This is the approach we adopt in this work.

We seek to train a PGML model that predicts the global annual energy yield potential ($Y$) of solar farms employing monofacial module technology (shown in Fig. 2). $Y$ is the maximal DC yield potential of a solar farm employing a given technology neglecting any time-dependent losses or limiting factors (e.g., module degradation, soiling, inverter losses). For a given farm design, the $Y$ can be computed with complex physics-based models that simulates optical, thermal, and electrical conditions the farm (10). The experience of developing such forward models enables us to immediately refine the input space and identify variables that are physically meaningful for a specific problem, i.e., determining $Y$. As mentioned, a small set of meteorological variables determine the $Y$ potential at a given location: the global horizontal irradiance ($I_{\text{GHI}}$), the ambient temperature ($T_{\text{amb}}$), and the clearness index ($k_t$) measuring the cloud conditions (i.e., direct vs. diffused sunlight contributions). It is easy to show that other features are either correlated (e.g., module temperature, $T_{\text{mod}}$) or redundant (e.g., longitude/latitude). It is important to note that the variables chosen should be physically significant for the target of interest. For instance, if the lifetime degradation of solar modules owing to solder bond failure were the metric of interest, then the range of temperature fluctuations ($\Delta T$) would be a physically-relevant variable to predict the phenomenon (11).

Having identified the physically significant variables, the problem of determining $Y$ can be cast as a regression problem. Let the monthly yield potential ($M$) be expressed as the function

$$M = f(I_{\text{GHI}}, T_{\text{amb}}, k_t) \qquad [1]$$

The desired quantity, $Y$ potential, is then readily calculated for each location by summing the predicted $M$, i.e., $Y = \sum_{i=1}^{12} M_i$. We train a simple generic fully-connected neural network to predict the the function in Eq. 1 (see Materials and Methods). Hereon, we will refer the neural network as the PGML model. A training dataset ($\mathcal{D}_i$) consists of $N$ monthly average values of $I_{\text{GHI}}$, $T_{\text{amb}}$, $k_t$, and the corresponding $M$ at different locations across the globe.

For the model to be generalizable to the globe, the training dataset needs to include samples that represent various combinations of input variables ($I_{\text{GHI}}$, $T_{\text{amb}}$, $k_t$) observed across different regions worldwide. Fortunately, these meteorological variables are often measured with high spatio-temporal resolution, thanks to the ever-expanding databases of satellite observations (12). In this study, we used NASA's POWER (13) climatological dataset as we focus on expected $Y$ at location (see Materials and Methods). However, it is unclear what quantity of observations ($N$) will be necessary to accurately approximate the function in Eq. 1 globally. Collecting a substantial amount of training data becomes impractical and cost-prohibitive because $M$ values must be either simulated or measured from fielded PV systems. To address this problem, we will exploit the problem-specific climatic equivalence among different regions on the planet to construct small, yet representative datasets.

**PV-specific climate zones enables diversity sampling.** Since ancient time, people have recognized that Earth's climate can be classified based on similarities in climatic characteristics (14). Several classification systems have been devised, e.g., Köppen–Geiger (K-G) (15), Thornthwaite (16), etc., each using different sets of variables to explain climatic similarity and variability across the planet. The set of variables are chosen based on their relevance to the issue at hand. For example, the popular K-G system uses precipitation and temperature to explain patterns in vegetation growth. Owing of its familiarity, K-G system has been extended to PV by incorporating irradiance (17). However, it considers variable that may only influence the yield indirectly: cloud condition in general plays a more direct role in determining PV performance compared to precipitation. Instead, by explicitly considering variables relevant to a problem, we create climate maps tailored to the problem in question. By doing so, we can identify the various regions in the world sharing similarities in climatic features that specifically influence PV energy yield. This also identifies the unique climes that need to be represented in a dataset if we are to train global scale model.

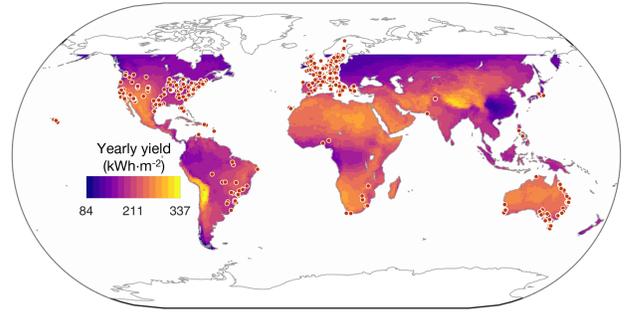

**Fig. 2.** Worldwide yearly yield potential of fixed-tilt, monofacial solar farms determined by worldwide gridded simulation in PVMAPS, a physics-based solar farm simulator. These results serve as the benchmark to test the data-efficient PGML developed in this paper through spatial diversity sampling. The red dots indicate the locations of over 3,000 PV systems with varying capacities, primarily concentrated in North America and Europe. Despite the large quantity of publicly available data, it is heterogeneous and lacks representation for certain PV zones (cf. Fig. 1(b)).

Here we introduce *PVZones* (PVZ), a climate map of PV energy data, that divides the world into seven PV energy yield specific zones ($z$) sharing similarity in $I_{\text{GHI}}$, $T_{\text{amb}}$ and $k_t$. Fig. 1(b) illustrates the seven geographical zones. The map was constructed by k-means ($k = 7$) algorithm to cluster annual average values of the three key features identified. We observe that zones are symmetric across the equator, forming latitudinally-arranged bands. This is expected given the Earth-Sun relationships, as each of the variables decreases as one moves away from the equator. We observe that larger countries (e.g., the United States and China) possess a greater diversity of climates as expected.

**Hypothesis: Sparse data can train a global scale ML model.** We hypothesize that by leveraging this climatic equivalence, the global annual yield potential could be determined from the data collected at a small number of strategically positioned test sites. This way such problem-specific climate map serves as a powerful tool for spatial sampling and guides the construction of diverse, representative datasets with a relatively few samples. To test the hypothesis, we evaluated two categories of datasets, representing the two extremes of controlled experiments. The first category comprises synthetic datasets where $M$ are estimated by a physics-based solar farm model, representing strictly controlled numerical experiments (shown in the top branch in Fig. 1). The second category comprises publicly sourced datasets with weak control and little to no oversight. The inherent and potentially unexplained heterogeneity of these datasets make them challenging to use them in training. This led us to develop a physics-based approach homogenize the dataset before they can be used for model training (see Fig. Fig. 1, bottom branch). Realistic data will have a degree of control within these two extremes. First, we demonstrate the validity of the hypothesis for the first (numerically computed synthetic data) category.

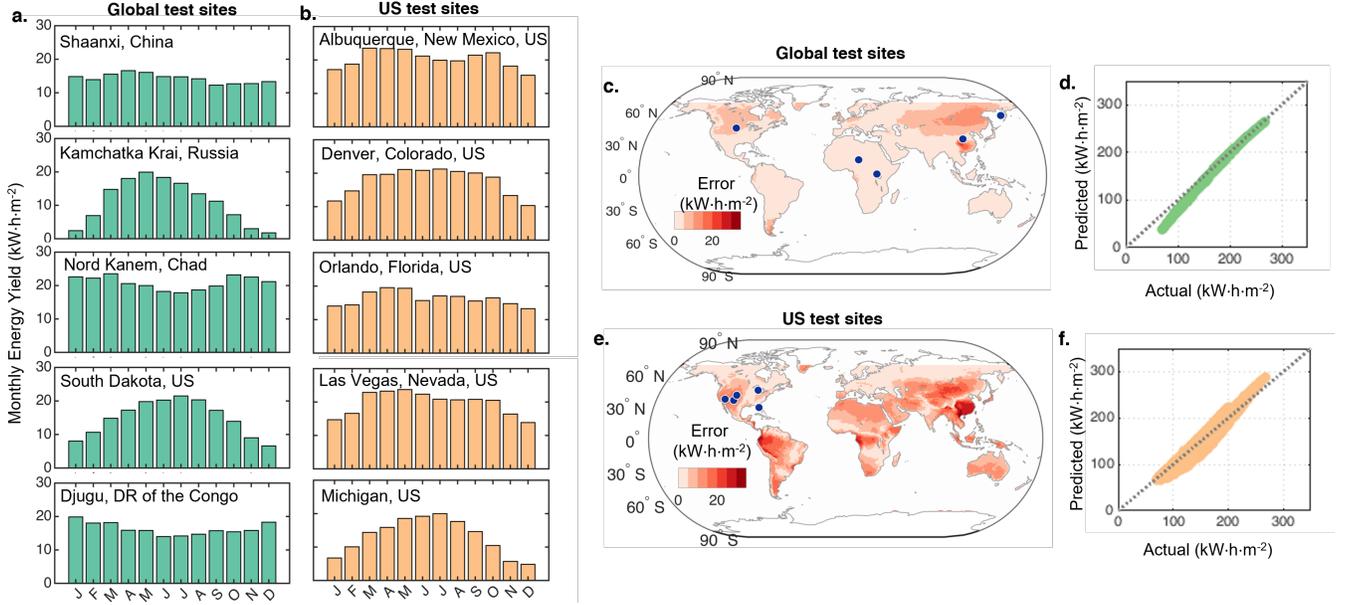

**Fig. 3.** Results with synthetic datasets. (a-b) Barplots showing the simulated monthly energy yields ($M$) used to construct the training data sets $\mathcal{S}_1$ and $\mathcal{S}_2$. The prediction accuracy of models trained on data from global data set (c-d) and DOE-NREL-Sandia's test site locations across US (e-f). Maps in (c) and (e) show that, with monthly yield data from 5 locations (blue dots), the models are able to predict yearly EY at most locations with less than just 16 kW h m$^{-2}$ error. (d) and (f) shows that the models are able to explain most of the variabilities.

### Synthetic datasets: strictly controlled numerical experiments

Two synthetic training datasets were created, each containing $M$ from five different locations. Dataset $\mathcal{S}_1$ contained data from five different sites in five different zones in PVZ. For dataset $\mathcal{S}_2$, the five selected sites were the locations of the US Department of Energy's (DOE) regional PV test centers located in the continental US. For each location, $M$ of monofacial solar farms were obtained by physics-based solar farm simulator PVMAPS. Figure 3(a-b) show the simulated $M$ at each test site. The simulated farms employed identical modules and had optimal fixed-tilt designs (see Materials and Methods). Both datasets contained only 60 observations of monthly average $I_{\text{GHI}}$, $T_{\text{amb}}$, $k_t$, and corresponding $M$.

**Predicting the global performance from five locations.** After training the PGML model with $\mathcal{S}_1$ and $\mathcal{S}_2$, $Y$ was computed for the globe. The model's ability to generalize was assessed by comparing the predicted $Y$ with the results obtained from a global physics-based simulation. Figures 3 (c-d) and (e-f) show the prediction accuracies of the models trained on $\mathcal{S}_1$ and $\mathcal{S}_2$, respectively. For the model trained on $\mathcal{S}_1$, we find that predictions for 95% locations are within 9 kW h m$^{-2}$ of the simulated $Y$, with higher errors typically occurring in zones that were not represented in the training dataset (e.g., Northern China, Mongolia). Fig. 3(d) shows that the model has good accuracy: the overall root mean squared error (RMSE) was only 3.85 kW h m$^{-2}$. Fig. 3(e-f) shows the prediction accuracy for the model trained on $\mathcal{S}_2$ dataset. Fig. 3(e) shows that, despite the sites being confined to the continental US, the model is still able to predict worldwide $Y$ potential for 95% of locations with less than 16 kW h m$^{-2}$ absolute error and a RMSE of 7.62 kW h m$^{-2}$. Again, the errors are higher primarily at locations not represented in the dataset. To see why $\mathcal{S}_2$ outperforms $\mathcal{S}_1$, we look at the parallel coordinates plots in Fig. 1(c-d) where each connected string represents an observation. $\mathcal{S}_1$ contains observations at wider range of $I_{\text{GHI}}$ and $T_{\text{amb}}$ compared to $\mathcal{S}_2$. These results are quite remarkable because they show that a diverse and high-quality experimental dataset from a small number of test sites can effectively train a ML model to accurately predict $Y$ potential across the world. Therefore, this results are consistent with our hypothesis for the first category of data.

### Heterogeneous crowd-sourced data: unsupervised experiments

Having demonstrated that a small, representative, high-quality synthetic dataset can predict the global YY potential, we now turn to utilizing already existing large, potentially low quality outdoor PV performance data. We collected monthly energy yields of over 3000 PV systems from various public PV performance data repositories like PVOutput, DuraMAT; online dashboards, and reports in the literature. Compared to the synthetic datasets, this data represent results from weakly controlled experiments. While such data may be available in large quantities, it is inherently heterogeneous. Although all systems employed monofacial modules, the systems varied in module attributes, design, and capacity. The nominal system capacities varied from a few kilowatts to megawatts (see SI Fig. S3). For many outdoor systems, parameters such as system capacity and tilt were unknown. The duration of the available data varied from just one year to multiple years.

**Correcting for heterogeneity.** Given the heterogenity, it is essential to verify whether the input variables can still adequately explain the output. For the heterogeneous systems, Eq. 1 can be reformulated as

$$M_{\text{field}} = f(I_{\text{GHI}}, T_{\text{amb}}, k_t, U) \quad [2]$$

where $U$ is a set of unknown variables that also influence the yield, e.g., system capacity, system orientation.

The primary source of variability in $Y_{\text{field}}$ is the variation in system capacities that leads to yield variations spanning four orders of magnitude. This disparity in system capacities can be corrected by scaling the measured $Y_{\text{field}}$ at a location to equate that of a reference PV system with known parameters at the same location. In the absence of existing physical reference systems, we

4 |

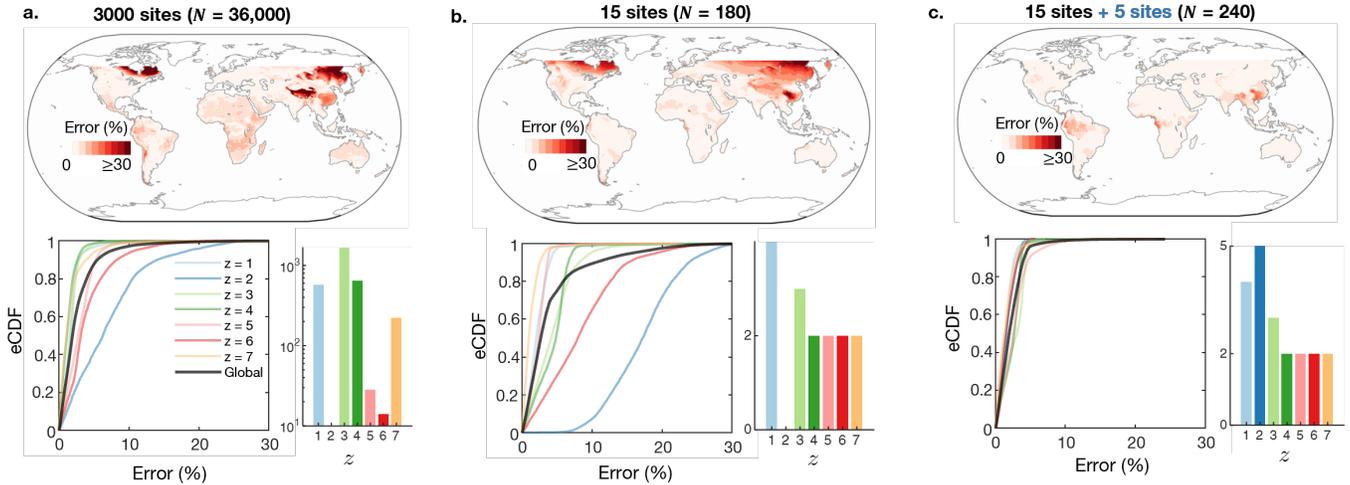

**Fig. 4.** Performance of the the PGML model trained on homogenized public PV performance data, corresponding zone-wise ($z$) cumulative distributions (eCDF) of error. The histograms show the number of sites in each zone for each training dataset. (a) When trained on the entire homogenized dataset $\mathcal{F}$, the error is less than 10% for 95% of the locations worldwide. Higher errors are observed at zones 2 and 6 due to their lack of representation in the dataset. (b) By choosing just 15 sites from $\mathcal{F}$ ensuring spatial diversity, similar accuracy can be achieved for all zones, except for zones 2 and 6. (c) By introducing synthetic data for 5 sites in zone 2, the error in these zones is significantly reduced resulting in less than 5% error for 95% of global locations.

simulated an optimally oriented system and use it as the reference ($Y_{\text{sim}}$). We assume that the $Y_{\text{field}}$ and the $Y_{\text{sim}}$ at a location is related by a scaling factor,

$$s = \frac{Y_{\text{field}}}{Y_{\text{sim}}} \quad [3]$$

$s$ is a constant coefficient for a given system that accounts for the difference in system capacity. After scaling, the monthly yields of the outdoor system become

$$M_{\text{field}}^* = s \times M_{\text{field}} \quad [4]$$

$MY_{\text{field}}^*$ can be seen as the output of the reference system perturbed by some noise, i.e.,

$$M_{\text{field}}^* = M_{\text{sim}} + \epsilon \quad [5]$$

The noise term ($\epsilon$) represents the discrepancy between $M_{\text{sim}}$ and $M_{\text{field}}^*$ owing to other unknown variables and model/measurement uncertainties. While $s$ corrects for the differences in the system capacities in the dataset, further reducing $\epsilon$ would require information about additional variables to explain the variability. Although algorithms exist to also correct for the tilt and azimuth parameters (18), it is reasonable to assume that the field systems were professionally installed and had near optimal orientation, i.e., tilt and azimuth, to maximize yield. Also, since some noise is inevitable in any outdoor measurement, such data represents a more practical scenario with weaker control over variability compared to the synthetic datasets discussed before.

After correcting for system capacity and computing the $M_{\text{field}}^*$ for all field locations, additional filtering were performed to remove the anomalous sites (see Materials and Methods). Finally, training dataset $\mathcal{F}$ ($N = 36,000$) was created to train the PGML model.

**Predicting the worldwide performance of solar farms.** Fig. 4 (a) shows the performance of the PGML trained on the entire homogenized dataset $\mathcal{F}$. With the large number of sites in this dataset we expect this model to perform well. The model demonstrates high accuracy consistent with this expectation: the prediction error was less than 10% compared to a PGML trained on global simulations able to predict $Y$ for 95% of the global sites.

However, we observe that the errors are higher at locations in zones 2 and 6 situated at higher latitudes. Such higher errors are expected since these two zones have the lowest number of observations in the training dataset, which is expected since zone 2 comprises regions near the Arctic Circle and the Himalayas that are largely inhabited. These higher latitude regions experience extreme fluctuations in temperature over the months, making them unique. This shows that even with large number of observations, the model may fail to generalize globally due to lack of representation for certain regions.

Next, we test our hypothesis that a smaller subset of this large dataset, *if diversely sampled*, could predict the $Y$ with a similar accuracy. For this, we chose 15 sites from $\mathcal{F}$ ensuring that each zone was represented, as shown in Fig. 4(b). We denote this smaller dataset by $\mathcal{F}_s$ ($N = 180$). Since no sites in $\mathcal{F}$ belonged to zone 2, it could not be represented. As the eCDF in Fig. 4(b) shows, despite having merely 0.5% of the observations of $\mathcal{F}$, the model achieves comparable accuracy (10% error 95th percentile) for zones other than 2 and 6. However, the overall accuracy of the global model is limited by the least represented zones, i.e., zones 2 and 6, as higher errors in these zones lead to an overall 15% error 95th percentile. Nevertheless, this result supports our hypothesis that a smaller dataset, when diversely sampled, contains sufficient information to make global scale yearly yield prediction.

We can improve this model by addressing the gaps in the representation for specific zones that evidently affects the accuracy of global prediction. By combining simulated results from 5 randomly chosen sites from zone 2 with $\mathcal{F}_s$, we constructed a fused dataset $\mathcal{F}_s^*$ ($N = 240$). As shown in Fig. 4(c), when trained on $\mathcal{F}_s^*$, the PGML outperformed $\mathcal{F}$ achieving an overall error of 5% (95th percentile). These results further support the hypothesis that data from sparse geographical locations can train a model to predict worldwide annual yields of solar farms.

**Validating against unseen field data.** To demonstrate the performance of PGML when trained on $\mathcal{F}$, we predict the $M_{\text{field}}^*$ at two locations. The sites are located in Ohio, USA ($z = 3$) (19) and Cochin, India ($z = 7$) (20). Although these data were not directly included in the training process, the zones were represented by different locations in these zones. Therefore, we expect the model to be able to generalize to these locations as well and predict the yearly yield performance. Indeed, for both locations the model is able to predict the field performance with good accuracy as shown in Fig. 5 and summarized in Table 1. For the site in Ohio, the error for $Y$ compared to $Y^*_{\text{field}}$ was 1.21 kW h m$^{-2}$ (0.9%), and the RMSE for $M^*_{\text{field}}$ predictions was 3.44 kW h m$^{-2}$ (10.8% MAPE). For the site in Cochin, the error for $Y$ was 0.64 kW h m$^{-2}$ (0.3%), and the RMSE for $M^*_{\text{field}}$ predictions was 0.66 kW h m$^{-2}$ (3.8% MAPE). Examples for numerous additional sites are provided in SI Sec. S5. These examples showcase the PGML's ability to generalize to arbitrary locations if their zone are represented in the homogenized training data.

**Table 1.** Comparison of the PGML predicted yields with simulated and field data.

| Sites | PVZone | MAPE $M_{\text{field}}$ | $Y_{\text{field}}$ | MAPE $M_{\text{sim}}$ |
|---|---|---|---|---|
| Ohio, USA | 3 | 10.8% | 0.9% | 12.8% |
| Cochin, India | 7 | 3.8% | 0.3% | 4.4% |

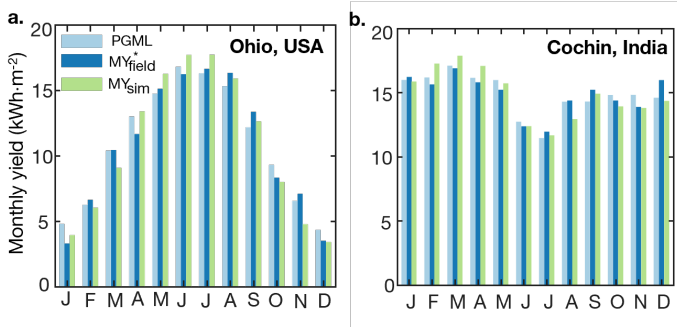

**Fig. 5.** Examples of $M$ and $Y$ predictions with the PGML model trained on public heterogenous data for farms at (a) Ohio, USA and (b) Cochin, India. The PGML predicts the $Y$ at these locations with just 0.9% and 2% relative errors compared to physics-based simulations ($Y_{\text{sim}}$) demonstrating the efficacy of the method when trained on homogenized, representative dataset.

## Discussion

The above results underscore the critical importance of data diversity in training the physics-based ML models. Since the PGML model essentially solves a regression problem, the generalization power of the model relies on how well a training dataset covers the gamut of the input domain. The representativeness of the different datasets can be visualized by the parallel plots shown in Fig. 1(c-d). The global input domain can be determined from the global gridded climatological datasets. By overlaying the observations of each training dataset, we can visually gauge the representativeness of the datasets and identify the potential range of values underrepresented in each dataset. It can be seen that $\mathcal{S}_1$, consisting of data from sites chosen randomly from different climate zones, has a better representation of the input domain compared to $\mathcal{S}_2$. The dataset built from field-performance data, $\mathcal{F}$, also has similar coverage to other datasets. However, $\mathcal{F}$ lacked representation for locations experiencing low monthly average temperatures ($<-5°C$), which explains the observed lower accuracy for zone 2. The fused dataset, $\mathcal{F}^*$, with the injection of simulation leads to significant improvements.

Recently, there has been a growing interest in data-centric AI models. Sampling diversely from variables physically-relevant variables eases the requirement for large quantity of data, which is expensive to obtain for any significant number of location while exercising good degree of experimental control. This way one may also reduce the number of parameters needed to learn the solution space, leading to smaller, data-efficient models. For this problem, the PGML model could be trained and inferences can be made in seconds on a modern personal computer. Due to their speed, such fast statistical model can complement more comprehensive detailed physics-based model to check for physical consistency, as demonstrated by their use in homogenization. The highly-efficient, lightweight, physics-based model can guide policy decisions for regional collaborations, determining the best sites for creating national/regional test centers, and cross-checking predictions provided by commercial and costly detailed analysis based on proprietary software.

In this paper, we have presented a method to assimilate global patterns in the energy yield potential of solar farms with data from strategically chosen locations. Achieving this requires training datasets that are representative of the domain of input variables. Towards this end, we have presented the concept of problem-specific climate maps to facilitate the construction of synoptic datasets. We have demonstrated that high-quality training data obtained a relatively few strategically determined test sites (e.g., five worldwide, if possible; or five DOE outdoor test facilities in the USA) can predict worldwide energy yield with very high accuracy. Further, valuable information can be deduced from publicly available PV performance data when system scale-dependent heterogeneity in the dataset is accounted for by rescaling to a reference system performance. The PGML model trained on yield potential data can be used to monitor the health of the PV systems overtime as the predictions represent the expected performance. However, attribution (degradation or change in system parameter) will require further analysis. The results leads to the following key conclusions.

1. Physical understanding should guide the choice of relevant variables for a PGML model. When predicting energy yield potential of a PV systems, the determining variables are $I_{\text{GHI}}$, $T_{\text{amb}}$ and $k_t$.

2. A large quantity of data alone is not sufficient to predict the $Y$ of PV system across the world. For accurate prediction across the globe, the dataset needs to be representative. We have shown that PV-specific climate maps (like PVZones) of physically relevant variables are a powerful tool to identify the different regions that needs to be represented to construct a representative dataset.

3. Since the methodology can be extended to any PV technologies and arbitrary farm topologies, strategically placed test sites provide a powerful empirical means to gauge the potential of the novel technologies. Thus, our results define an innovative data-driven, physics-based machine learning model for creating resources to accelerate the worldwide adoption of PV technologies.

Further, the approach is not limited to the problem of predicting solar farm performance but can be adapted for a broad range of problems where the input space of physically significant variables can be partitioned, analyzed and exploited.

## Materials and Methods

**Climate data.** Climatic normal dataset containing the monthly averages of observations since 1984 was obtained from NASA POWER. The global dataset has a spatial resolution of $0.5°\times0.5°$ and consists of data for over 250,000 locations.

**Constructing PVZones for energy yield.** The zones in the PVZones map (shown in Fig. 1(b)) were determined by k-means clustering of physically relevant variables. To predict energy yields, the primary variables are $I_{\text{GHI}}$, $T_{\text{amb}}$, and $k_t$. The clustering was performed with the annual average values of these variables.

**PGML model.** The machine-learning model for yield prediction is a feed-forward neural network consisting of two hidden layers (10 neurons each). The network was implemented and trained with MATLAB's Deep Learning toolbox. During training, Bayesian regularization was used to obtain the optimal number of parameters while penalizing over-fitting.

**Global physics-based solar farm simulation.** To test and validate the machine learning model, we used the PVMAPS to simulate the reference worldwide yearly yield potential of monofacial solar farms. Various aspects of the physical model is described in detail in ref. (10). We simulated systems employing monofacial modules ($\eta$ = 24%) installed in a fixed tilt configuration. During the simulation, the array pitch ($p$) was assumed to be 3 m and the module height ($h$) and elevation were assumed 1 m. The modules were optimally tilted to maximize the annual power output. The physical simulation is performed on a coarse grid with a spatial resolution of $2° \times 18°$ (1200 locations) on monthly basis for a year with climatological normal data. Finally, the reference yearly yields (shown in Fig. 2) were calculated using the method described in SI Sec. S1.

**Global model error.** The error of PGML model is computed by comparing the predicting the $Y$ on a regular $0.5° \times 0.5°$ grid using the following equation.

$$E = \frac{|Y_{\text{pred}} - Y^*_{\text{pred}}|}{Y^*_{\text{pred}}} \times 100 \qquad [6]$$

$Y_{\text{pred}}$ is the yield predicted by PGML model and $Y^*_{\text{pred}}$ is the yearly yield predicted by a reference PGML trained on gridded simulation data. $Y^*_{\text{pred}}$ is shown in SI Fig. S1.

**Public PV performance data.** Outdoor PV performance data for monofacial PV systems are collected from public dashboards and literature from over 3000 different PV systems around the globe (see Fig. S2). If available, $M_{\text{field}}$ data from multiple years were averaged. Raw $M_{\text{field}}$ data were filtered using the procedure described in SI Fig. S4. First, raw $M_{\text{field}}$ data was scaled to with factor $s$ to obtain $M^*_{\text{field}}$ (in kW h m$^{-2}$) as described in the main paper. Then, RMSE was computed with respect to the monthly yields predicted by the physics-based simulation ($M_{\text{sim}}$) for the same location. Finally, sites with RMSE greater than the threshold $e_{\text{th}}$ (marked red) were removed. $e_{\text{th}}$ was set to 3 standard deviations of the overall RMSE distribution.

**ACKNOWLEDGMENTS.** We acknowledge the insightful discussion with Henry DeRyke (Purdue University, USA), Dr. Reza Asadpour (Purdue University, USA), Professor Mohammad Ryyan Khan (East West University, Bangladesh), and Professor Peter Bermel (Purdue University, USA).

# Supporting Information for

Physics-guided machine learning predicts the planet-scale performance of solar farms with sparse, heterogeneous, public data

Jabir Bin Jahangir and Muhammad A. Alam

Corresponding author: Muhammad A. Alam
Email: alam@purdue.edu

**This PDF file includes:**

Figures S1 to S7
Table S1

## S1. Global solar farm yield potential

The global predictive ability of the PGML models was compared with energy yield potential computed with a physics-based simulator run on a global grid. However, such simulations require significant computational time and resources, making it impractical for global simulations with high spatial resolution. A naïve interpolation of the yearly energy yield (Y) from coarse-gridded simulation does not account for the location-specific spatial variation of the input variables, hence can be inaccurate at intermediate points. Fortunately, a PGML model trained on coarse-gridded simulations can be used for *functional* interpolation as it considers the spatial variation of input conditions (1).

The reference global worldwide potential was thus determined in two steps. First, global solar farm monthly (M) and yearly (Y) energy yield potentials for fixed-tilted monofacial modules were calculated using PVMAPS (a physics-based solar farm simulator, (2) for details) for a coarse grid of 2° × 18°. The simulation parameters are shown in Table S1. The computed yield potentials are shown in Fig. S1(a). Then, these results from physics-based simulation are used to determine the yield potentials with higher spatial resolution (0.5° × 0.5°) as shown in Fig. S1 (b). This is achieved by training the PGML model on $\mathcal{S}_{\text{world}}$ containing the results of coarse-gridded simulation. The $\mathcal{S}_{\text{world}}$ dataset ($N = 14400$) contained $I_{\text{GHI}}, T_{\text{amb}}, k_t$ as inputs and corresponding simulated monthly yields as outputs. The goodness of fit of the trained PGML model when predicting for monthly (M) and yearly (Y) yields are shown in Fig. S1 (c-d), respectively. In both cases, predicted vs. actual plots show that the PGML model is able to explain the variability in the data accurately.

**Table S1. PVMAPS simulation parameters used to generate data for $\mathcal{S}_{\text{world}}$**

| Parameter | Value |
|---|---|
| Module Type | Monofacial |
| Module efficiency (η) | 24% |
| Module height | 1 m |
| Farm topology | Fixed-tilt |
| Tilt | Optimal* |
| Azimuth | Optimal* |
| Row-to-row spacing | 3 m |

* Location-specific optimal tilt and azimuth maximizing the energy yield

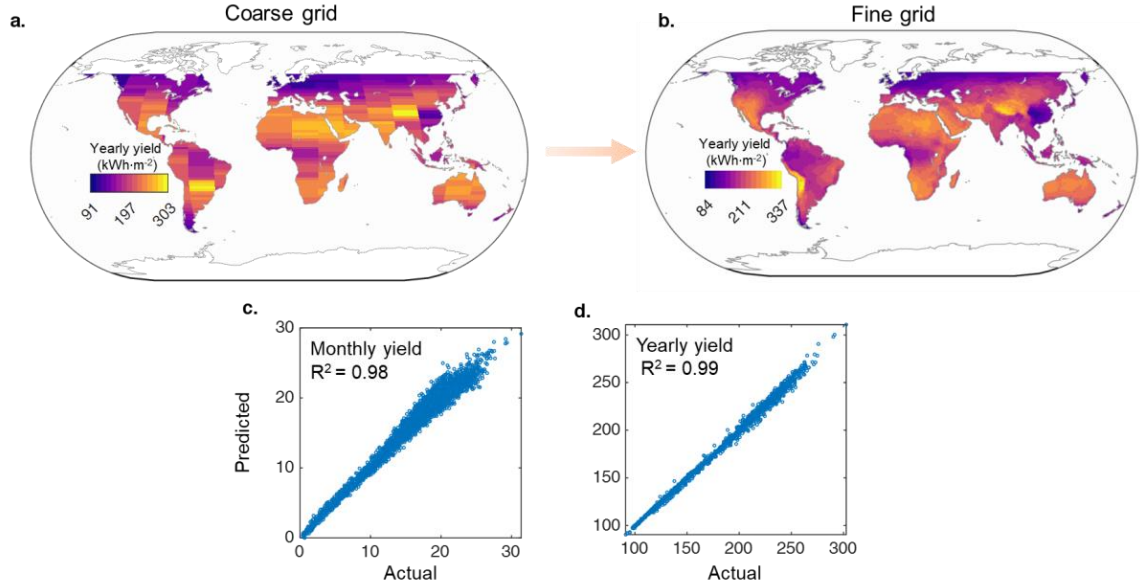

**Fig. S1** Global yearly yield potential of fixed-tilt solar farms on (a) coarse grid (2° × 18°) and (b) fine grid (0.5° × 0.5°). A PGML trained on the coarse-gridded simulation results is used to compute the yields on the finer grid. This PGML model predicts the (c) monthly and (d) yearly yield with high coefficient of determination ($R^2$).

## S2. Heterogenous dataset

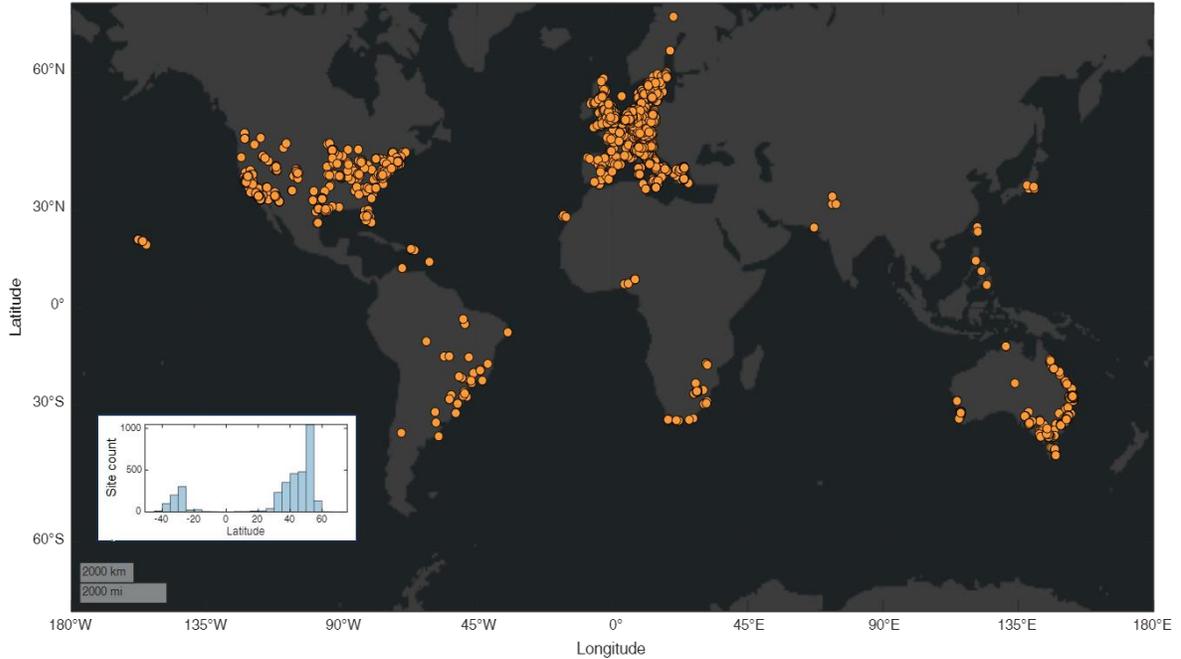

**Fig. S2** Locations of PV sites in the dataset $\mathcal{F}$ and their latitudinal distribution. Monthly yield from each location was collected for at least one year.

**Capacity distribution and resulting variability in yield.** For the data collected in this dataset, the capacity of PV systems varied between a few kW to MW as shown in Fig. S3(a). The yearly yield (YY in kWh) is highly correlated with system capacity, leading to YY variation over 3 orders of magnitude (Fig. S3 (b)). Since for some systems the capacity was unknown or misreported, it could not be readily used as an independent variable.

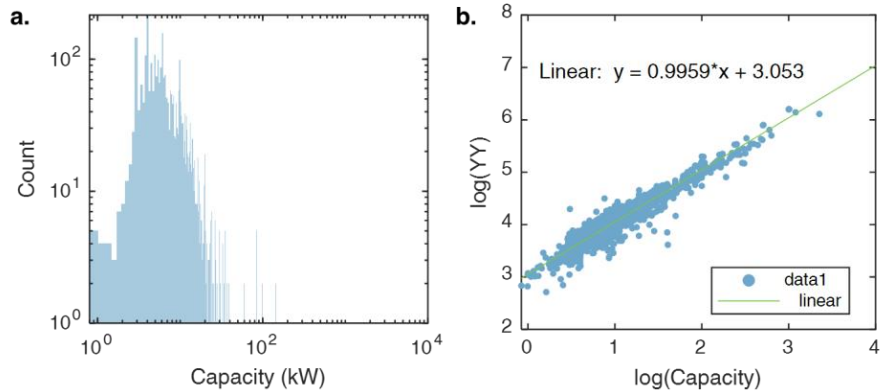

**Fig. S3** (a) Distribution of known site capacities in the dataset $\mathcal{F}$ and (b) correlation between the site capacity and reported yearly yield.

**Filtering anomalous monthly profiles.** Some of these systems are unsupervised, and the numbers reported may not be reliable. The dataset even includes systems other than PV (e.g., heater). To filter the monthly yield profiles associated with those systems, we use the PGML model described in Sec. S1 to produce the expected monthly yield profiles at a given location. The procedure is shown in Fig. S4. Profiles resulting in large RMSE are removed, as the yield-numbers being physically impossible for the specific location. This physics-guided detection and removal of outliers in a key component of the overall algorithm.

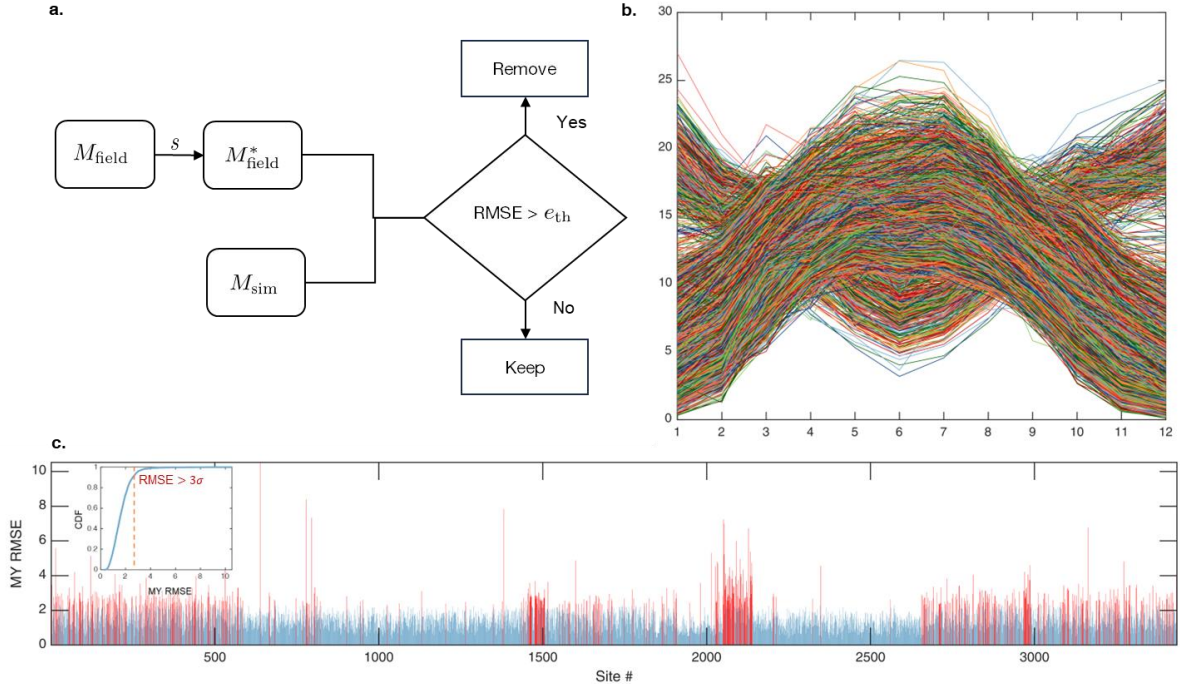

**Fig. S4** (a) Procedure for filtering sites with anomalous monthly yield (MY) profile. First, raw $M_{field}$ (in kWh·m$^{-2}$) data is scaled to with factor $s$ to obtain $M^*_{field}$ (b) as described in the main paper. Then, RMSE is computed with respect to the monthly yields predicted by the physics-based simulation ($M_{sim}$) for the same location. (c) Sites with RMSE greater than the error threshold $e_{th}$ (marked red) are removed. $e_{th}$ was set to 3 standard deviations of the overall MY RMSE distribution.

## S4. Diversity sampling vs. random sampling from large dataset $\mathcal{F}$.

To illustrate the impact of data quantity ($N$), we trained PGMLs with increasing number of sites randomly chosen from $\mathcal{F}$. The performance of these models is shown in Fig. S5 (a-d). The global yearly yield prediction accuracy of the PGML model increases with the increasing representation of different PV zones. Fig. S5(a) shows that when the model was trained on data from a single location ($N = 12$) in zone 5, the model is only able to make accurate predictions for zone 5 and its adjacent zones, but the accuracy suffers in zones 2 and 6. The cumulative distribution (eCDF) shows that that at 95% sites the errors lie within 20%. When a PGML is trained on data from 15 sites ($N = 180$) as shown in Fig. S5(b), the overall errors do not improve significantly and remains especially higher for sites in zone 2 that lacks representation (see Fig. 3(a)). When compared to the result shown in Fig. 3(b) where 15 sites were also picked ensuring each zone is represented, the higher error obtained here demonstrates the efficacy of spatial diversity sampling.

The lack of representation for zone 2 in the geographic distribution of sites becomes more evident when the model is trained with all 3000 sites ($N = 36000$) as shown in Fig. S5(c) (also in Fig. 3(a)). Although the overall accuracy of the model improves significantly ($\leq$ 10% error at 95% locations), we still find significantly higher errors in zone 2 which limits overall model accuracy. This underscores the view that a large quantity of data, if not representative, will not be sufficient for a global scale model. Now, this gap in the data can be remedied by introducing simulated data for zone 2 in $\mathcal{F}$. This way, we constructed an additional dataset $\mathcal{F}^*(N = 37200)$ by fusing simulated (100 sites in zone 2) and field dataset $\mathcal{F}$. The resultant model shows impressive accuracy (Fig. S5(c)) with the errors at 95% locations lying below just 5%. With spatial diversity sampling, as demonstrated in the main paper, similar accuracy for yearly yield prediction can be obtained with only 20 ($N = 240$) sites.

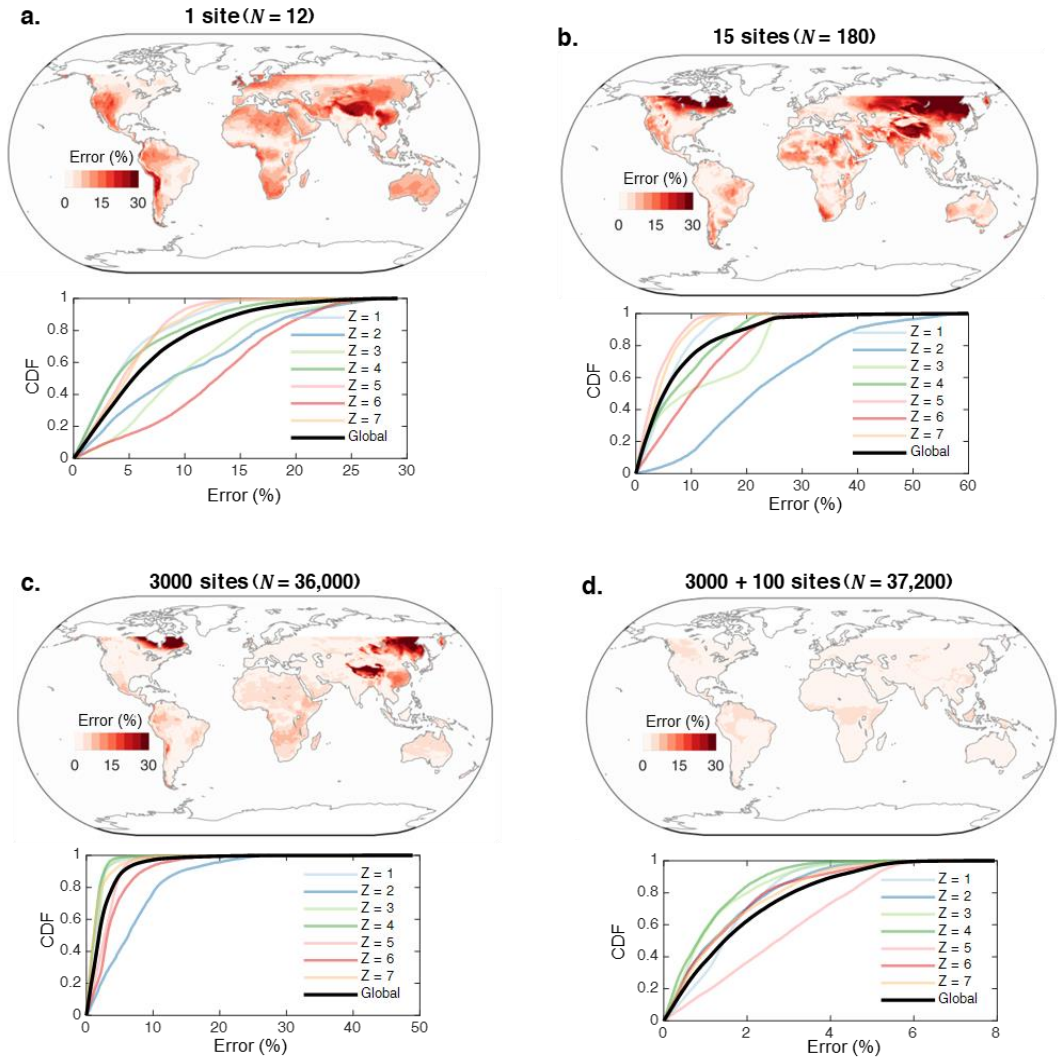

**Fig. S5** Performance of PGML model when (a) 1 and (b) 15 sites are randomly chosen from $\mathcal{F}$. The higher errors in (b) compared to Fig. 3(b) shows that randomly choosing a small subset does not necessarily obtain high accuracy global model. (c) While the PGML trained on the complete dataset (3000 sites) shows reduced errors overall, higher errors are still observed for zones 2 due to the lack of representation. (d) By including additional synthetic data from 100 sites in zone 2, the errors are significantly reduced.

**S5. Predicting monthly yield profiles for unseen sites**

Fig. S6 compares the monthly yield profiles predicted by various models for a number of sites not included in the training dataset. Images of some of these systems are shown in Fig. S7. $M_{\mathrm{PGML}}$ denotes the yield predicted by the PGML model trained on dataset $\mathcal{F}^*(N = 37200)$ mentioned above. $M^*_{\mathrm{field}}$ and $M_{\mathrm{sim}}$ denote the scaled field data and PVMAPS simulated yield values, respectively. Ideally, we want these quantities to be as close to each other as possible. Looking at the examples, we notice that field collected homogenized monthly yields $M^*_{\mathrm{field}}$ do not always agree with $M_{\mathrm{PGML}}$ and even $M_{\mathrm{sim}}$ at every location. However, $M_{\mathrm{PGML}}$ and $M_{\mathrm{sim}}$ agree in nearly all locations during most months.

Several factors contribute to the discrepancy ($\epsilon$) beyond the inherent error associated with machine learning. We have assumed that the system parameters (e.g., capacity, orientation) of the systems remained constant throughout the data collection period. In reality, the active capacity could vary over months due to maintenance and repair, or simply faulty instruments, which would result in a significant change in the monthly yield profiles. Another source of error is the mismatch between actual inputs and the climatic average values assumed during training. Since the data was available for a year for many of these locations, the input values could vary from the climatic average. However, such variabilities in the system parameters and inputs can be accounted for in coordinated controlled experiments.

Nevertheless, the strong agreement between $M_{\mathrm{PGML}}$ and $M_{\mathrm{sim}}$ indicates that the PGML model can be robust to these variabilities with large enough observations from each zone. Since public field data is naturally susceptible to these variabilities, increasing the number of observations may help improve the accuracy of monthly yield predictions.

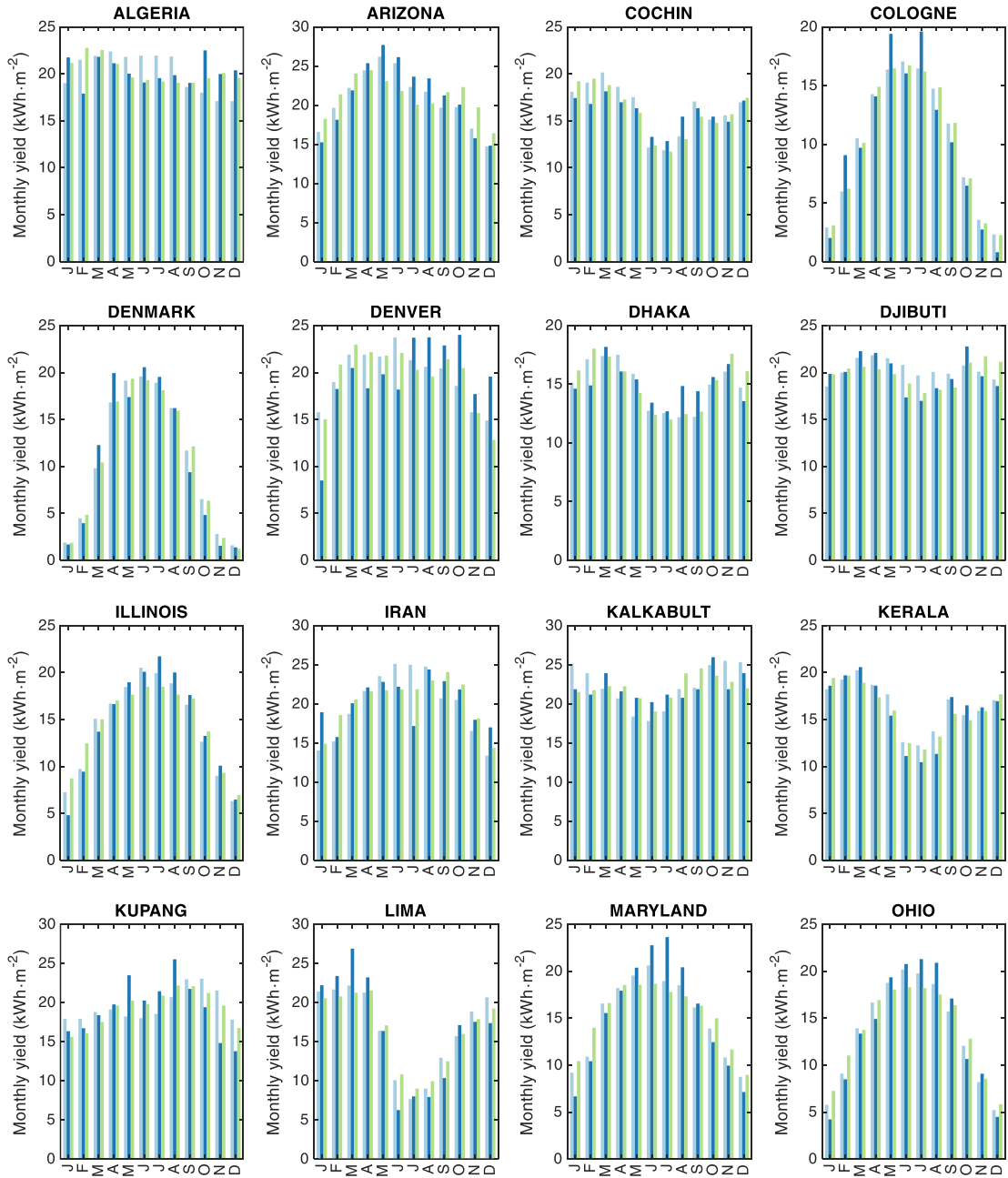

**Fig. S6** Monthly yield profiles for different sites predicted by PGML trained on the fused dataset $\mathcal{F}^*(N = 37200)$. Despite trained on field data, the PGML predicts yields consistent with $M_{sim}$.

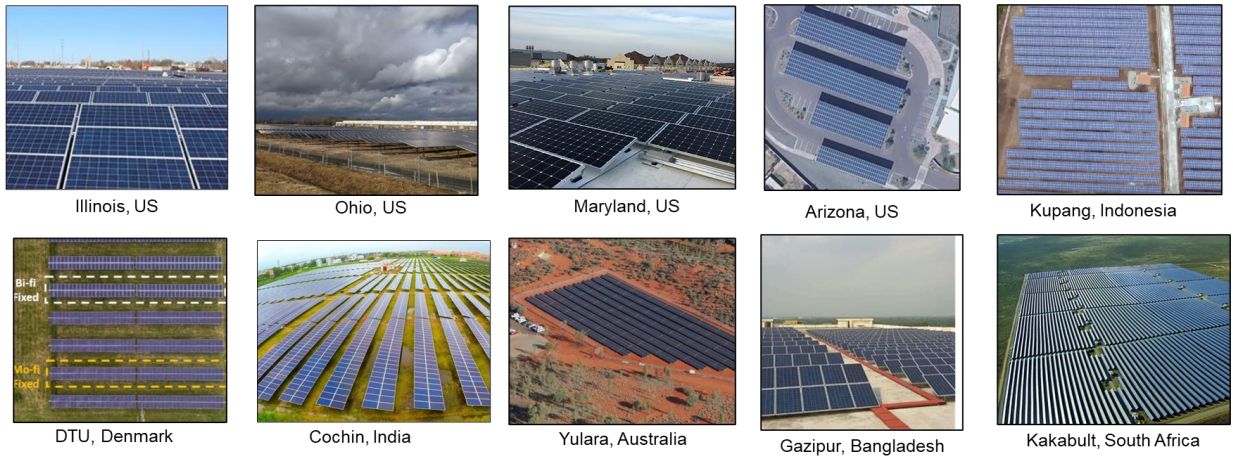

**Fig. S7** Images of some of the diverse PV systems for which energy yields are being predicted.